\def\year{2017}\relax
\documentclass[letterpaper]{article}
\usepackage{aaai17}
\usepackage{times}
\usepackage{helvet}
\usepackage{courier}
\usepackage{graphicx}
\usepackage{latexsym}
\usepackage{amsmath}
\usepackage{hyperref}
\begin{document}
% The file aaai.sty is the style file for AAAI Press 
% proceedings, working notes, and technical reports.
%
\title{SummaRuNNer: A Recurrent Neural Network based Sequence Model for Extractive Summarization of Documents}
\author{Ramesh Nallapati, Feifei Zhai\thanks{Work was done while the author was an employee at IBM.}, Bowen Zhou\\
nallapati@us.ibm.com,ffzhai2012@gmail.com, zhou@us.ibm.com\\
1011 Kitchawan Road, Yorktown Heights, NY 10598
}
\maketitle
\begin{abstract}
We present {\it SummaRuNNer}, a Recurrent Neural Network (RNN) based sequence model for extractive summarization of documents and show that it achieves  performance better than or comparable to state-of-the-art. Our model has the additional advantage of being very interpretable, since it allows visualization of its predictions broken up by abstract features such as information content, salience and novelty. Another novel contribution of our work is abstractive training of our extractive model that can train  on human generated reference summaries alone, eliminating the need for sentence-level extractive labels.
\end{abstract}

\section{Introduction}
Document summarization is an important problem that has many applications in information retrieval and natural language understanding. Summarization techniques are mainly classified into two categories: extractive and abstractive. Extractive methods aim to select salient snippets, sentences or passages from documents, while abstractive summarization techniques aim to concisely paraphrase the information content in the documents.

A vast majority of the literature on document summarization is devoted to extractive summarization. Traditional methods for extractive summarization can be broadly classified into greedy approaches ({\it e.g.}, \cite{carbonell:98}), graph based approaches ({\it e.g.}, \cite{erkan:04}) and constraint optimization based approaches ({\it e.g.}, \cite{McDonald:07}). 

Recently, neural network based approaches have become popular for extractive summarization.  For example,
\cite{Kageback:14} employed the recursive autoencoder \cite{Socher:11} to summarize documents, producing best performance on the Opinosis dataset \cite{ganesan:10}. \cite{yin:15} applied Convolutional Neural Networks (CNN) to project sentences to continuous vector space and then select sentences by minimizing the cost based on their `prestige' and `diverseness', on the task of multi-document extractive summarization. Another related work is that of \cite{cao:16}, who address the problem of query-focused multi-document summarization using CNNs, where they use weighted-sum pooling over sentence representations to represent documents. The weights are learned from attention over sentence representations based on the query. 

%algorithm that greedily selects sentences based on relevance and redundancy.
%Taking a different approach, \cite{McDonald:07} formulated sentence selection as a constraint optimization problem where ILP was used to minimize the cost incurred, in terms of relevance and redundancy. 
%Graph based approaches like LexRank~ and Markov Random walk~\cite{wan:08}, on the other hand, assign edge weights as the lexical similarity between sentences represented as nodes in the graph, and use variants of PageRank~\cite{Page:99} to discover cluster centroids which are salient sentences in the document.

%The work by  ~\cite{Kobayashi:15} demonstrated that it is reasonably good to use the similarity between the sentence embedding and document embedding for saliency measurement, where the document embedding is derived from sum pooling of sentence embeddings. 
%Further developing this idea, 

Recently, with the emergence of strong generative neural models for text \cite{bahdanau:14}, abstractive techniques are also becoming increasingly popular. For example, \cite{rush:15} proposed an attentional feed-forward network for abstractive summarization of sentences into short headlines. Further developing on their work, \cite{nallapati} propose a set of recurrent neural network based encoder-decoder models that focus on various aspects of summarization like handling out-of-vocabulary words and modeling syntactic features of words in the sentence. In a follow-up work \cite{nallapati_conll}, they also propose abstractive techniques for summarization of large documents into multi-sentence summaries, using the CNN/DailyMail corpus\footnote{https://github.com/deepmind/rc-data}.

Despite the emergence of abstractive techniques, extractive techniques are still attractive as they are less complex, less expensive, and generate grammatically and semantically correct summaries most of the time. In a very recent work, Cheng and Lapata \shortcite{jianpeng} proposed an attentional encoder-decoder for extractive single-document summarization and applied to the CNN/Daily Mail corpus.  
%Unlike most previous approaches for extractive summarization, their model is trained on a very large dataset obtained from a news source called Daily Mail, which was originally constructed by \cite{reading_comprehension} for passage-based question answering, and also used for abstractive summarization in \cite{nallapati}. 

Like \cite{jianpeng}, our work also focuses only on {\it sentential} extractive summarization of single documents using neural networks. We use the same corpus used by \cite{nallapati_conll} and \cite{jianpeng} for our experiments, since its large size makes it attractive for training deep neural networks such as ours, with several thousands of parameters. 

Our main contributions are as follows: (a) we propose SummaRuNNer, a simple recurrent network based sequence classifier that outperforms or matches state-of-the-art models for extractive summarization; (b) the simple formulation of our model facilitates interpretable visualization of its decisions; and (c) we present a novel training mechanism that allows our extractive model to be trained end-to-end using abstractive summaries.  
%Our model is much simpler than the one proposed in \cite{jianpeng}, yet we show it produces superior performance when trained on the same Dail Mail corpus. Further, we propose a simple technique to optimally convert abstractive gold summaries into extractive labels without add
%technique which uses hierarchical attention-based encoder-decoder architecture. First, they perform a using an RNN to select sentences. Then, they generate an optimal ordering of subset of words from the extracted sentences. \\

\section{SummaRuNNer}
In this work, we treat extractive summarization as a sequence classification problem wherein, each sentence is visited sequentially in the original document order and a binary decision is made (taking into account previous decisions made) in terms of whether or not it should be included in the summary. We use a GRU based Recurrent Neural Network \cite{gru_rnn} as the basic building block of our sequence classifier. A GRU-RNN is a recurrent network with two gates, ${\bf u}$ called the update gate and ${\bf r}$ , the reset gate, and can be described by the following equations:

\begin{eqnarray}
{\bf u}_j &=&  \sigma({\bf W}_{ux}{\bf x}_j + {\bf W}_{uh} {\bf h}_{j-1} + {\bf b}_u) \label{eq:update_gate}\\
{\bf r}_j &=&  \sigma({\bf W}_{rx}{\bf x}_j + {\bf W}_{rh} {\bf h}_{j-1} + {\bf b}_r) \label{eq:reset_gate}\\
{\bf h}'_j &=& \tanh({\bf W}_{hx}{\bf x}_j + {\bf W}_{hh} ({\bf r}_j \odot {\bf h}_{j-1}) + {\bf b}_h) \label{eq:hidden_state}\\
{\bf h}_j &=& (1-{\bf u}_j)\odot {\bf h}'_j + {\bf u}_j \odot {\bf h}_{j-1}
\end{eqnarray}
where the ${\bf W}$'s and ${\bf b}$'s are the parameters of the GRU-RNN and ${\bf h}_j$ is the real-valued hidden-state vector at timestep $j$ and ${\bf x}_j$ is the corresponding input vector, and $\odot$ represents the Hadamard product.

\begin{figure}[ht]
    \vspace{-0.3in}
	\centering
  \includegraphics[width=0.55\textwidth]{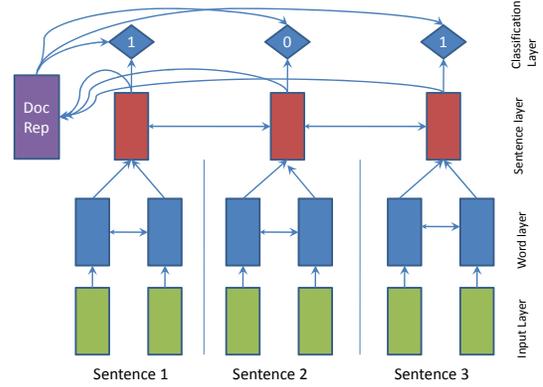}
  	\vspace{-0.5in}
	\caption{{\small SummaRuNNer: A two-layer RNN based sequence classifier: the bottom layer operates at word level within each sentence, while the top layer runs over sentences. Double-pointed arrows indicate a bi-directional RNN. The top layer with 1's and 0's is the sigmoid activation based classification layer that decides whether or not each sentence belongs to the summary. The decision at each sentence depends on the content richness of the sentence, its salience with respect to the document, its novelty with respect to the accumulated summary representation and other positional features.}}
	\label{fig:rnn_classifier}
\end{figure}

Our model consists of a two-layer bi-directional GRU-RNN, whose graphical representation is presented in Figure \ref{fig:rnn_classifier}. 
The first layer of the RNN runs at the word level, and computes hidden state representations at each word position sequentially, based on the current word embeddings and the previous hidden state. %, as shown by the following equations.
We also use another RNN at the word level that runs backwards from the last word to the first, and we refer to the pair of forward and backward RNNs as a bi-directional RNN. The model also consists of a second layer of bi-directional RNN that runs at the sentence-level and accepts the average-pooled, concatenated hidden states of the bi-directional word-level RNNs as input. The hidden states of the second layer RNN encode the representations of the sentences in the document. The representation of the entire document is then modeled as a non-linear transformation of the average pooling of the concatenated hidden states of the  bi-directional sentence-level RNN, as shown below.
\begin{equation}
{\bf d} = \tanh(W_d\frac{1}{N_d}\sum_{j=1}^{N^d} [{\bf h}^{f}_j, {\bf h}^{b}_j] + {\bf b}),
\end{equation}
where ${\bf h}^f_j$ and ${\bf h}^b_j$ are the hidden states corresponding to the $j^{th}$ sentence of the forward and backward sentence-level RNNs respectively, $N_d$ is the number of sentences in the document and `$[ ]$' represents vector concatenation. 

For classification, each sentence is revisited sequentially in a second pass, where a logistic layer makes a binary decision as to whether that sentence belongs to the summary, as shown below.
\begin{eqnarray}\label{eq:classification}
P(y_j=1|{\bf h}_j, {\bf s}_j, {\bf d}) = \sigma(W_{c}{\bf h}_j && \verb|#(content)| \nonumber\\
                                      + {\bf h}_j^TW_{s}{\bf d} && \verb|#(salience)| \nonumber\\
 - {\bf h}_j^TW_{r}\tanh({\bf s_j}) && \verb|#(novelty)| \nonumber\\ 
 + W_{ap}{\bf p}^a_j && \verb|#(abs. pos. imp.)| \nonumber\\
 + W_{rp}{\bf p}^r_j  && \verb|#(rel. pos. imp.)| \nonumber\\
 + b),&& \verb|#(bias term)| 
\end{eqnarray}
where $y_j$ is a binary variable indicating whether the $j^{th}$ sentence is part of the summary, ${\bf h}_j$, the representation of the sentence is given by a non-linear transformation of the concatenated hidden states at the $j^{th}$ time step of the bidirectional sentence-level RNN, and ${\bf s}_j$ is the dynamic representation of the summary at the $j^{th}$ sentence position, given by:
\begin{equation}\label{eq:summary_rep}
{\bf s}_j = \sum_{i=1}^{j-1}{\bf h}_iP(y_i=1|{\bf h}_i, {\bf s}_i, {\bf d}).
\end{equation}
In other words, the summary representation is simply a running weighted summation of all the sentence-level hidden states visited till sentence $j$, where the weights are given by their respective probabilities of summary membership.
%\begin{equation}
%{\bf h}_i = \tanh(W[{\bf h}^{f}_i, {\bf h}^{b}_i]+{\bf b}),
%\end{equation}

In Eqn. (\ref{eq:classification}), the term $W_c{\bf h}_j$ represents the information content of the $j^{th}$ sentence, ${\bf h}_j^TW_{s}{\bf d}$ denotes the salience of the sentence with respect to the document, ${\bf h}_j^TW_{r}\tanh({\bf s_j})$ captures the redundancy of the sentence with respect to the current state of the summary\footnote{We squash the summary representation using the $\tanh$ operation so that the magnitude of summary remains the same for all time-steps.}, while the next two terms model the notion of the importance of the absolute and relative position of the sentence with respect to the document.\footnote{The absolute position denotes the actual sentence number, whereas the relative position refers to a quantized representation that divides each document into a fixed number of segments and computes the segment ID of a given sentence.} We consider ${\bf p}^a$ and ${\bf p}^r$, the absolute and relative positional embeddings respectively, as model parameters as well.

We minimize the negative log-likelihood of the observed labels at training time.
\begin{eqnarray}
l({\bf W}, {\bf b}) &=& -\sum_{d=1}^N \sum_{j=1}^{N_d}(y_j^d\log P(y_j^d=1|{\bf h}_j^d,{\bf s}_j^d,{\bf d}_d) \nonumber\\
& +& (1-y_j^d)\log(1- P(y_j^d=1|{\bf h}_j^d,{\bf s}_j^d,{\bf d}_d))\nonumber\\
\label{eq:lhood}
\end{eqnarray}
where ${\bf x}$ is the document representation and ${\bf y}$ is the vector of its binary summary labels. 
At test time, the model emits probability of summary membership $P(y_j)$ at each sentence sequentially, which is used as the model's soft prediction of the extractive summary. 

\subsection{Extractive Training}\label{sec:extractive_training}
 In order to train our extractive model, we need ground truth in the form of sentence-level binary labels for each document, representing their membership in the summary. However, most summarization corpora only contain human written abstractive summaries as ground truth. To solve this problem, we use an unsupervised approach to convert the abstractive summaries to extractive labels. Our approach is based on the idea that the selected sentences from the document should be the ones that maximize the Rouge score with respect to gold summaries. Since it is computationally expensive to find a globally optimal subset of sentences that maximizes the Rouge score, we employ a greedy approach, where we add one sentence at a time incrementally to the summary, such that the Rouge score of the current set of selected sentences is maximized with respect to the entire gold summary . 
%Since Rouge-2 and Rouge-L are order sensitive, we respect the original ordering of the selected sentences in the document while computing Rouge scores. 
We stop when none of the remaining candidate sentences improves the Rouge score upon addition to the current summary set. We return this subset of sentences as the extractive ground-truth, which is used to train our RNN based sequence classifier. 

\subsection{\bf Abstractive Training}\label{sec:abstractive_training}
In this section, we propose a novel training technique to train SummaRuNNer abstractively, thus eliminating the need to generate approximate extractive labels. To train SummaRuNNer using reference summaries, we couple it with an RNN decoder that models the generation of abstractive summaries at training time only. The RNN decoder uses the summary representation at the last time-step of SummaRuNNer as context, which modifies Eqs. \ref{eq:update_gate} through \ref{eq:hidden_state} as follows:
\begin{eqnarray}
{\bf u}_k &=&  \sigma({\bf W'}_{ux}{\bf x}_k + {\bf W'}_{uh} {\bf h}_{k-1} + {\bf W'}_{uc}{\bf s}_{-1} + {\bf b'}_u) \nonumber\\
{\bf r}_k &=&  \sigma({\bf W'}_{rx}{\bf x}_k + {\bf W'}_{rh} {\bf h}_{k-1} + {\bf W'}_{rc}{\bf s}_{-1} + {\bf b'}_r) \nonumber\\
{\bf h}'_k &=& \tanh({\bf W'}_{hx}{\bf x}_k + {\bf W'}_{hh} ({\bf r}_k \odot {\bf h}_{k-1}) + \nonumber\\
 && {\bf W'}_{hc}{\bf s}_{-1} + {\bf b'}_h)\nonumber
\end{eqnarray}
where ${\bf s}_{-1}$ is the summary representation as computed at the last sentence of the sentence-level bidirectional RNN of SummaRuNNer as shown in Eq. \ref{eq:summary_rep}. The parameters of the decoder are distinguished from those of SummaRuNNer using the `prime' notation, and the time-steps of the decoder use index $k$ to distinguish word positions in the summary from sentence indices $j$ in the original document. For each time-step of the decoder, the embedding of the word from the previous time-step is treated as its input ${\bf x}_k$.

Further, the decoder is equipped with a soft-max layer to emit a word at each time-step. The emission at each time-step is determined by a feed-forward layer $f$ followed by a softmax layer that assigns ${\bf p}_k$, probabilities over the entire vocabulary at each time-step, as shown below.
\begin{eqnarray}
{\bf f}_k &=& \tanh({\bf W'}_{fh}{\bf h}_k + {\bf W'}_{fx}{\bf x}_k + {\bf W'}_{fc}{\bf s}_{-1}+ {\bf b'}_f) \nonumber\\
{\bf P_v(w)}_k &=& \mbox{softmax}({\bf W'}_v{\bf f}_k + {\bf b'}_v) \nonumber
\end{eqnarray}
Instead of optimizing the log-likelihood of the extractive ground truth as shown in Eq. \ref{eq:lhood}, we minimize the negative log-likelihood of the words in the reference summary as follows.
\begin{equation}
l({\bf W}, {\bf b}, {\bf W'}, {\bf b'}) = -\sum_{k=1}^{N_s} \log ({\bf P_v}(w_k))
\end{equation}
where $N_s$ is the number of words in the reference summary. At test time, we uncouple the decoder from SummaRuNNer and emit only the sentence-level extractive probabilities ${\bf p(y_j)}$ of Eq. \ref{eq:classification}. 

Intuitively, since the summary representation ${\bf s}_{-1}$ acts as the only information channel between the SummaRuNNer model and the decoder, maximizing the probability of abstractive summary words as computed by the decoder will require the model to learn a good summary representation which in turn depends on accurate estimates of extractive probabilities ${\bf p(y_j)}$. %Therefore, we hope that our end-to-end abstractive training leads the model to learn accurate extractive probabilities naturally.  

\section{Related Work}

Treating document summarization as a sequence classification model has been considered by earlier researchers. For example, \cite{crf4summarization} used Conditional Random Fields to binary-classify sentences sequentially. Our approach is different from theirs in the sense that we use RNNs in our model that do not require any handcrafted features for representing sentences and documents. 

Since the sequence classifier requires sentence-level summary membership labels to train on, we used a simple greedy approach to convert the abstractive summaries to extractive labels. Similar approaches have been employed by other researchers such as \cite{svore}. Further, recently \cite{tgsum} propose an ILP based approach to solve this problem optimally. 

Most single-document summarization datasets available for research such as DUC corpora are not large enough to train deep learning models. Two recent papers (\cite{nallapati_conll} and \cite{jianpeng}) solve this problem by proposing a new corpus based on news stories from CNN and Daily Mail that consist of around 280,000 documents and human generated summaries. Of these, the work of \cite{jianpeng} is the closest to our work since they also employ an extractive approach for summarization. Their model is based on an encoder-decoder approach where the encoder learns the representation of sentences and documents while the decoder classifies each sentence based on encoder's representations using an attention mechanism. Our model, when extractively trained, employs a single sequence model with no decoder, and therefore may have fewer parameters. Our abstractively trained model has a decoder too, but it is different from that of \cite{jianpeng} since our decoder is used to model the likelihood of abstractive gold summaries at training time, so as to eliminate the need for extractive labels. Their model, on the other hand, requires extractive labels even with the decoder. In fact, unlike our unsupervised greedy approach to convert abstractive summaries to extractive labels, \cite{jianpeng} chose to train a separate supervised classifier using manually created labels on a subset of the data. This may yield more accurate gold extractive labels, but incurs additional annotation costs.

The work of \cite{nallapati_conll} also uses an encoder-decoder approach, but is fully abstractive in the sense that it generates its own summaries at test time. Our abstractive trainer comes close to their work, but only generates sentence-extraction probabilities at test time. We include comparison numbers with this work too, in the following section.

\section{Experiments and Results}\label{sec:exp}

\begin{figure*}[htpb]
    \vspace{-0.6in}
	\centering
  \includegraphics[width=\textwidth]{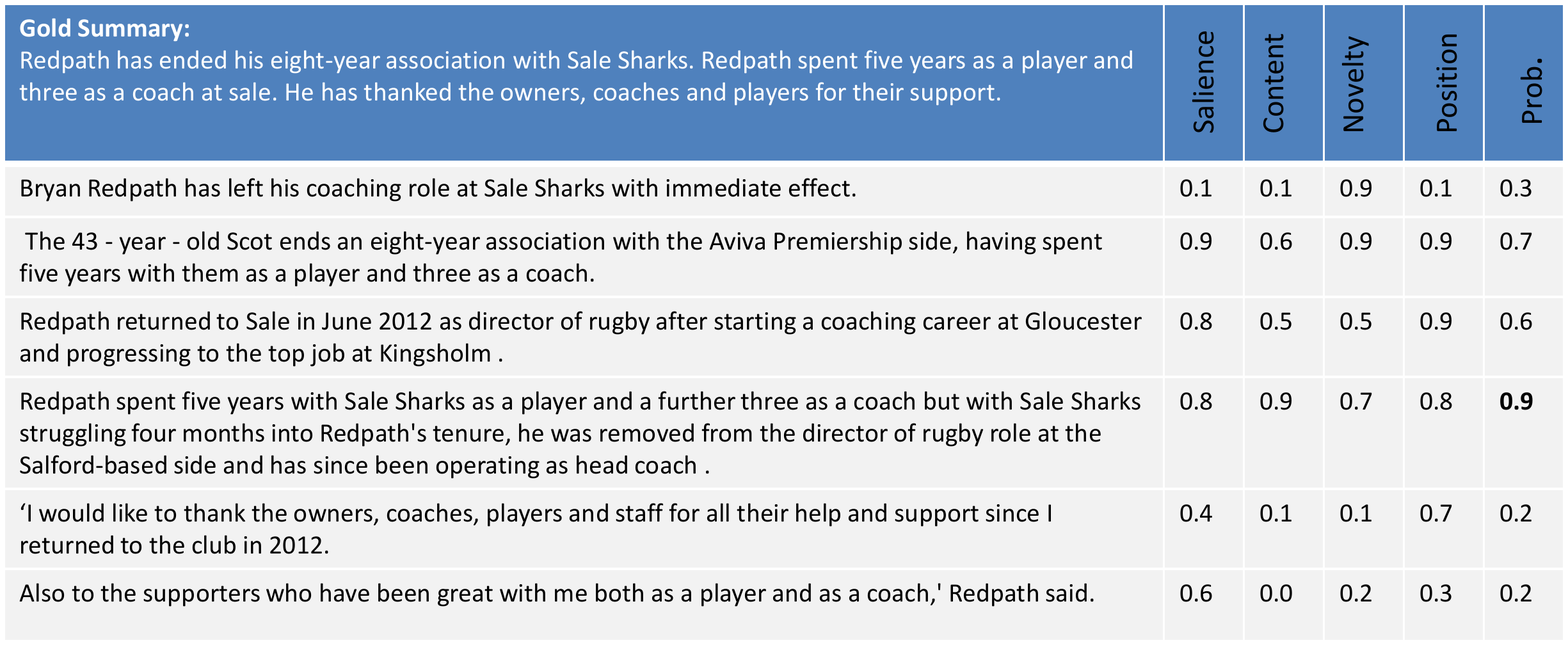}
  	\vspace{-2.7in}
	\caption{{\small Visualization of SummaRuNNer output on a representative document. Each row is a sentence in the document, while the shading-color intensity is proportional to its probability of being in the summary, as estimated by the RNN-based sequence classifier. In the columns are the normalized scores from each of the abstract features in Eqn. (\ref{eq:classification}) as well as the final prediction probability (last column). Sentence 2 is estimated to be the most salient, while the longest one, sentence 4, is considered the most content-rich, and not surprisingly, the first sentence the most novel. The third sentence gets the best position based score.}}
	\label{fig:heatmap}
\end{figure*}

\subsection{Corpora} 
For our experiments, we used the CNN/DailyMail corpus originally constructed by \cite{reading_comprehension} for the task of passage-based question answering, and re-purposed for the task of document summarization as proposed in   \cite{jianpeng} for extractive summarization and \cite{nallapati_conll} for abstractive summarization. In order to make a fair comparison with the former, we left out the CNN subset of the corpus, as done by them. To compare with the latter, we used the joint CNN/Daily Mail corpora. Overall, we have 196,557 training documents, 12,147 validation documents and 10,396 test documents from the Daily Mail corpus. If we also include the CNN subset, we have 286,722 training documents, 13,362 validation documents and 11,480 test documents. On average, there are about 28 sentences per document in the training set, and an average of 3-4 sentences in the reference summaries. The average word count per document in the training set is 802.

We also used the DUC 2002 single-document summarization dataset\footnote{http://www-nlpir.nist.gov/projects/duc/guidelines/2002.html} consisting of 567 documents as an additional out-of-domain test set to evaluate our models.
%We report performance of our model trained only on the Daily Mail corpus as done by \cite{jianpeng} as well as one trained on a combination of CNN and DailyMail sources. %We also report the Rouge scores of the pseudo-ground truth as the upperbound performance for our model.

\subsection{Evaluation} 
In our experiments below, we evaluate the performance of SummaRuNNer using different variants of the Rouge metric \footnote{{\small\url{ http://www.berouge.com/Pages/default.aspx}}} computed with respect to the gold summaries. To compare with \cite{jianpeng} on the Daily Mail corpus, we use limited length Rouge recall and 75 bytes and 275 bytes as reported by them. To compare with \cite{nallapati_conll} on the CNN/Daily Mail corpus, we use the same full-length Rouge F1 metric used by the authors. On DUC 2002 corpus, following the official guidelines, we use the limited length Rouge recall metric at  75 words. We report the scores from Rouge-1, Rouge-2 and Rouge-L, which are computed using the matches of unigrams, bigrams and longest common subsequences respectively, with the ground truth summaries.

\subsection{Baselines} 
On all datasets, we use Lead-3 model, which simply produces the leading three sentences of the document as the summary as a baseline. On the Daily Mail and DUC 2002 corpora, we also report performance of LReg, a feature-rich logistic classifier used as a baseline by \cite{jianpeng}. On DUC 2002 corpus, we report several baselines such as Integer Linear Programming based approach \cite{ilp}, and graph based approaches such as TGRAPH \cite{tgraph} and URANK \cite{urank} which achieve very high performance on this corpus. In addition, we also compare with the state-of-the art deep learning models from \cite{jianpeng} and \cite{nallapati_conll}.

\subsection{SummaRuNNer Settings} We used 100-dimensional {\it word2vec} \cite{mikolovWord2vec:13} embeddings trained on the CNN/Daily Mail corpus as our embedding initialization. We limited the vocabulary size to 150K and the maximum number of sentences per document to 100, and the maximum sentence length to 50 words, to speed up computation. We fixed the model hidden state size at 200. We used a batch size of 64 at training time, and {\it adadelta} \cite{adadelta} to train our model. We employed gradient clipping to regularize our model and an early stopping criterion based on validation cost. We trained SummaRuNNer both extractively as well as abstractively. When the model is abstractively trained, we denote it as {\it SummaRuNNer-abs} in the results.
%Our model typically converges in 5 epochs on this dataset.

At test time, picking all sentences with $P(y=1)\ge 0.5$ may not be an optimal strategy since the training data is very imbalanced in terms of summary-membership of sentences. Instead, we pick sentences sorted by the predicted probabilites until we exceed the length limit when limited-length Rouge is used for evaluation. When full-length F1 is used as the metric, we fixed the number of top sentences to be selected based on the validation set. 

\subsection{Results on Daily Mail corpus}

Table \ref{tab:test_perf_75b} shows the performance comparison of SummaRuNNer with state-of-the-art model of \cite{jianpeng} and other baselines on the DailyMail corpus using Rouge recall with summary length restricted to 75 bytes. While the abstractively trained SummaRuNNer performs on par with the state-of-the-art model, the extractively trained model significantly improves over their model.

%on the test set using Rouge scores with respect to the gold summaries. The results, reported  in Table \ref{tab:test_perf}, show that our model significantly outperforms the state-of-the-art system of \cite{jianpeng} and several other baselines, on all Rouge variants. 
%We also outperform with statistical significance another simple, but very strong baseline that simply outputs the first 75 words of the document as the summary\footnote{since 75 words is the cut-off limit for Rouge recall metric}. Since most of the salient content in news stories is found in the first few sentences, this is a very hard baseline to beat in this domain. %The performance improves marginally when we also add the CNN dataset to train the model. 

\begin{table}[ht]
\centering
\begin{tabular}{|l|l|l|l|}
\hline
 & Rouge-1 & Rouge-2 & Rouge-L \\
 \hline
Lead-3  & 21.9 & 7.2 & 11.6 \\
LReg(500) & 18.5 & 6.9 & 10.2 \\
Cheng {\it et al} '16 & 22.7 & 8.5 & 12.5 \\
SummaRuNNer-abs & 23.8 & 9.6 & 13.3 \\
SummaRuNNer &  {\bf 26.2}$\pm 0.4$*  & {\bf 10.8}$\pm 0.3$* & {\bf 14.4}$\pm 0.3$* \\
\hline
\end{tabular}
\caption{{\small Performance of various models on the {\bf entire Daily Mail test set} using the {\bf limited length recall} variants of Rouge with respect to the abstractive ground truth at {\bf 75 bytes}.  Entries with asterisk are statistically significant using 95\% confidence interval with respect to the nearest model, as estimated by the Rouge script.}}
\label{tab:test_perf_75b}
\end{table}

In Table \ref{tab:test_perf_275b}, we report the performance of our model with respect to Rouge recall at 275 bytes of summary length. In this case, our abstractively trained model underperforms the extractive model  of \cite{jianpeng} while the extractively trained model is statistically indistinguishable from their model. This shows that the SummaRuNNer is better at picking the best sentence for summarization than the subsequent ones.

\begin{table}[ht]
\centering
\begin{tabular}{|l|l|l|l|}
\hline
 & Rouge-1 & Rouge-2 & Rouge-L \\
 \hline
Lead-3 & 40.5 & 14.9 & 32.6 \\
Cheng {\it et al} '16 & {\bf 42.2} & {\bf 17.3} & {\bf 34.8}* \\
SummaRuNNer-abs &  40.4  & 15.5  & 32.0 \\
SummaRuNNer &  42.0 $\pm 0.2$ & 16.9 $\pm 0.4$  & 34.1 $\pm 0.3$ \\
\hline
\end{tabular}
\caption{{\small Performance of various models on the {\bf entire Daily Mail test set} using the {\bf limited length recall} variants of Rouge at {\bf 275 bytes}. SummaRuNNer is statistically indistinguishable from the model of  Cheng et al, `16 at 95\% C.I. on Rouge-1 and Rouge-2.}}
\label{tab:test_perf_275b}
\end{table}

One potential reason SummaRuNNer does not consistently outperform the extractive model of \cite{jianpeng} is the additional supervised training they used to create sentence-level extractive labels to train their model. Our model instead uses an unsupervised greedy approximation to create extractive labels from abstractive summaries, and as a result, may be more noisy than their ground truth.

\subsection{Results on CNN/Daily Mail corpus}

We also report the performance of SummaRuNNer on the joint CNN/Daily Mail corpus. The only other work that reports performance on this dataset is the abstractive encoder-decoder based model of \cite{nallapati_conll}, in which they use full-length F1 as the metric since neural abstractive approaches can learn when to stop generating words in the summary. In order to do a fair comparison with their work, we use the same metric as them. On this dataset, SummaRuNNer significantly outperforms their model as shown in Table \ref{tab:perf_cnndailymail}. The superior performance of our model is not entirely surprising since abstractive summarization is a much harder problem, but the table serves to quantify the current performance gap between extractive and abstractive approaches to summarization. The results also demonstrate the difficulty of using the F1 metric for extractive summarization since SummaRuNNer, with its top three sentences with highest prediction probability as the summary, errs on the side of high recall at the expense of precision. Dynamically adjusting the summary length based on predicted probability distribution may help balance precision and recall and may further boost F1 performance, but we have not experimented with it in this work.

\begin{table}[ht]
\centering
\begin{tabular}{|l|l|l|l|}
\hline
 & Rouge-1 & Rouge-2 & Rouge-L \\
 \hline
 Lead-3 & 39.2 & 15.7 & {\bf 35.5} \\
\cite{nallapati_conll} & 35.4 & 13.3 & 32.6  \\
SummaRuNNer-abs & 37.5 & 14.5 & 33.4 \\
SummaRuNNer & {\bf 39.6}$\pm0.2$* & {\bf 16.2}$\pm0.2$* & 35.3$\pm0.2$ \\
\hline
\end{tabular}
\caption{{\small Performance comparison of abstractive and extractive models on the entire CNN Daily Mail test set using {\bf full-length F1} variants of Rouge. SummaRuNNer is able to significantly outperform the abstractive state-of-the-art as well as the Lead-3 baseline (on Rouge-1 and Rouge-2).}}
\label{tab:perf_cnndailymail}
\end{table}

\subsection{Results on the Out-of-Domain DUC 2002 corpus}

We also evaluated the models trained on the DailyMail corpus on the out-of-domain DUC 2002 set as shown in Table \ref{tab:perf_duc2002}. SummaRuNNer is again statistically on par with the model of \cite{jianpeng}. However, both models perform worse than graph-based TGRAPH \cite{tgraph} and URANK \cite{urank} algorithms, which are the state-of-the-art models on this corpus.  Deep learning based supervised models such as SummaRuNNer and that of \cite{jianpeng} perform very well on the domain they are trained on, but may suffer from domain adaptation issues when tested on a different corpus such as DUC 2002. Graph based unsupervised approaches, on the other hand, may be more robust to domain variations.

\begin{table}[ht]
\centering
\begin{tabular}{|l|l|l|l|}
\hline
 & Rouge-1 & Rouge-2 & Rouge-L \\
 \hline
Lead-3  & 43.6 & 21.0 & 40.2 \\
LReg & 43.8 & 20.7 & 40.3 \\
ILP & 45.4 & 21.3 & 42.8 \\
TGRAPH & 48.1 & {\bf 24.3}* & - \\
URANK & {\bf 48.5}* & 21.5 & - \\
Cheng {\it et al} '16 & 47.4 & 23.0 & {\bf 43.5} \\
SummaRuNNer-abs & 44.8 & 21.0 & 41.2 \\
SummaRuNNer &  46.6 $\pm 0.8$  & 23.1 $\pm 0.9$ & 43.03 $\pm 0.8$ \\
\hline
\end{tabular}
\caption{{\small Performance of various models on the {\bf DUC 2002} set using the {\bf limited length recall} variants of Rouge at {\bf 75 words}. SummaRuNNer is statistically within the margin of error at 95\% C.I. with respect to Cheng et al `16, but both are lower than state-of-the-art results.}}
\label{tab:perf_duc2002}
\end{table}

\section{Qualitative Analysis}
In addition to being a state-of-the-art performer, SummaRuNNer has the additional advantage of being very interpretable. The clearly separated terms in the classification layer (see Eqn. \ref{eq:classification}) allow us to tease out various factors responsible for the classification of each sentence. This is illustrated in Figure \ref{fig:heatmap}, where we display a representative document from our validation set along with normalized scores from each abstract feature responsible for its final classification. Such visualization is especially useful in explaining to the end-user the decisions made by the system.

We also display a couple of example documents from the Daily Mail and DUC corpora highlighting the sentences chosen by SummaRuNNer and comparing them with the gold summary in Table \ref{tab:example_summaries}. The examples demonstrate qualitatively that SummaRuNNer performs a reasonably good job in identifying the key points of the document.

\section{Conclusion}
In this work, we propose a very interpretable neural sequence model for extractive document summarization that allows intuitive visualization, and show that it is better performing than or is comparable to the state-of-the-art deep learning models. 

We also propose a novel abstractive training mechanism to eliminate the need for extractive labels at training time, but this approach is still a couple of Rouge points below our extractive training on most datasets. We plan to further explore combining extractive and abstractive approaches as part of our future work. One simple approach could be to pre-train the extractive model using abstractive training. Further, we plan to construct a joint extractive-abstractive model where the predictions of our extractive component form stochastic intermediate units to be consumed by the abstractive component.

\begin{table}
\centering
{\small
\begin{tabular}{|p{8.5cm}|}
\hline
{\it Document:}  {\bf @entity0 have an interest in @entity3 defender @entity2 but are unlikely to make a move until january} .  {\bf the 00 - year - old @entity6 captain has yet to open talks over a new contract at @entity3 and his current deal runs out in 0000 .}  @entity3 defender @entity2 could be targeted by @entity0 in the january transfer window @entity0 like @entity2 but do n't expect @entity3 to sell yet they know he will be free to talk to foreign clubs from january .  @entity12 will make a £ 0million offer for @entity3 goalkeeper @entity14 this summer . the 00 - year - old is poised to leave @entity16 and wants to play for a @entity18 contender . {\bf @entity12 are set to make a £ 0million bid for @entity2 's @entity3 team - mate @entity14 in the summer }   \\
\hline
{\it Gold Summary:} @entity2 's contract at @entity3 expires at the end of next season . 00 - year - old has yet to open talks over a new deal at @entity16 . @entity14 is poised to leave @entity3 at the end of the season \\
\hline
\hline
{\it Document:} {\bf today , the foreign ministry said that control operations carried out by the corvette spiro against a korean-flagged as received ship fishing illegally in argentine waters were carried out `` in accordance with international law and in coordination with the foreign ministry '' .} the foreign ministry thus approved the intervention by the argentine corvette when it discovered the korean ship chin yuan hsing violating argentine jurisdictional waters on 00 may . ...  {\bf the korean ship , which had been fishing illegally in argentine waters , was sunk by its own crew after failing to answer to the argentine ship 's warnings .} the crew was transferred to the chin chuan hsing , which was sailing nearby and approached to rescue the crew of the sinking ship .....\\
\hline
{\it Gold Summary:} the korean-flagged fishing vessel chin yuan hsing was scuttled in waters off argentina on 00 may 0000 . adverse weather conditions prevailed when the argentine corvette spiro spotted the korean ship fishing illegally in restricted argentine waters . the korean vessel did not respond to the corvette 's warning . instead , the korean crew sank their ship , and transferred to another korean ship sailing nearby . in accordance with a uk-argentine agreement , the argentine navy turned the surveillance of the second korean vessel over to the british when it approached within 00 nautical miles of the malvinas ( falkland ) islands . \\
\hline
\end{tabular}
}
\caption{{\small Example documents and gold summaries from Daily Mail (top) and DUC 2002 (bottom) corpora. The sentences chosen by SummaRuNNer for extractive summarization are highlighted in bold.}}
\label{tab:example_summaries}
\end{table}

%\pagebreak
\bibliographystyle{aaai}
\bibliography{references}

\end{document}